\documentclass[review]{elsarticle}

\usepackage{hyperref}

\journal{Pattern Recognition}

\usepackage{graphicx}
\usepackage{comment}
\usepackage{amsmath,amssymb} 
\usepackage{color}
\usepackage{gensymb}
\usepackage{algorithm, algorithmic}
\usepackage{multirow}
\usepackage{pdfpages}
\usepackage{array}
\usepackage{booktabs}
\usepackage{pgfplotstable}

\newcommand{\etal}{{\em et al\,.}}       
\newcommand{\eg}{{\em e.g.}}           
\newcommand{\ie}{{\em i.e.}}           
\newcommand{\aka}{{\em a.k.a.}}           

\begin{document}

\begin{frontmatter}

\title{Generalizable Model-agnostic Semantic Segmentation via Target-specific Normalization}
\author[nju]{Jian Zhang} 
\author[seu]{Lei Qi\corref{cor}}
\author[nju]{Yinghuan Shi\corref{cor}}
\author[nju]{Yang Gao}

\cortext[cor]{Corresponding author: Yinghuan Shi (syh@nju.edu.cn) and Lei Qi (qilei@seu.edu.cn)}

\address[nju]{State Key Laboratory for Novel Software Technology, Nanjing University, China}
\address[seu]{Key Lab of Computer Network and Information Integration (Ministry of Education), Southeast University, China.}

\begin{abstract}
    Semantic segmentation in a supervised learning manner has achieved significant progress in recent years. However, its performance usually drops dramatically due to the data-distribution discrepancy between seen and unseen domains when we directly deploy the trained model to segment the images of unseen (or new coming) domains. To this end, we propose a novel domain generalization framework for the generalizable semantic segmentation task, which enhances the generalization ability of the model from two different views, {including the training paradigm and the test strategy}. Concretely, we exploit the model-agnostic learning to simulate the domain shift problem, which deals with the domain generalization from the training scheme perspective. Besides, considering the data-distribution discrepancy between seen source and unseen target domains, we develop the target-specific normalization scheme to enhance the generalization ability. Furthermore, when images come one by one in the test stage, we design the image-based memory bank (Image Bank in short) with style-based selection policy to select similar images to obtain more accurate statistics of normalization. Extensive experiments highlight that the proposed method produces state-of-the-art performance for the domain generalization of semantic segmentation on multiple benchmark segmentation datasets, \ie, Cityscapes, Mapillary. 
    
    \end{abstract}


\end{frontmatter}


\section{Introduction}

Semantic segmentation is a long-standing yet challenging task in computer vision {and pattern recognition} community, {which aims to} assign the semantic label {(\eg, person or car)} to each pixel in {a given} image. {Currently,} in real-world tasks, semantic segmentation has a wide range of applications, such as autonomous driving, security {surveillance}, and augmented reality. 
Recently, {thanks to deep convolutional neural networks,} we have witnessed the significant breakthrough of semantic segmentation on various benchmark datasets.
Those powerful methods perform well in {an identical distribution assumption between training and test images in the traditional supervised setting} -- the training and test images are collected from the same domain. Despite their success, when this assumption could not satisfy in the real case, \ie, when the training and test images come from different domains, the performance drops dramatically.

{Therefore, unsupervised domain adaptation~(UDA)~\cite{chen2019crdoco,chang2019all,sun2019not} is proposed to deal with this issue. In the setting of UDA, the model was trained with 1) labeled data from the source domain and 2) unlabeled data from the target domain. Basically, the common goal of UDA methods is to mitigate the domain shift between the source domain and the target domain. In this meaning, the model is expected to generalize well in the target domain. However, when applying the trained UDA segmentation model to other unseen domains, it still requires to first collect unlabeled data from new domains and then retrain the model with the newly collected data. In this case, the burden from two sides, \ie, data collection and model retraining, is inevitable.}

\begin{figure}
  \centering  
  \begin{center}
  \includegraphics[scale=0.2]{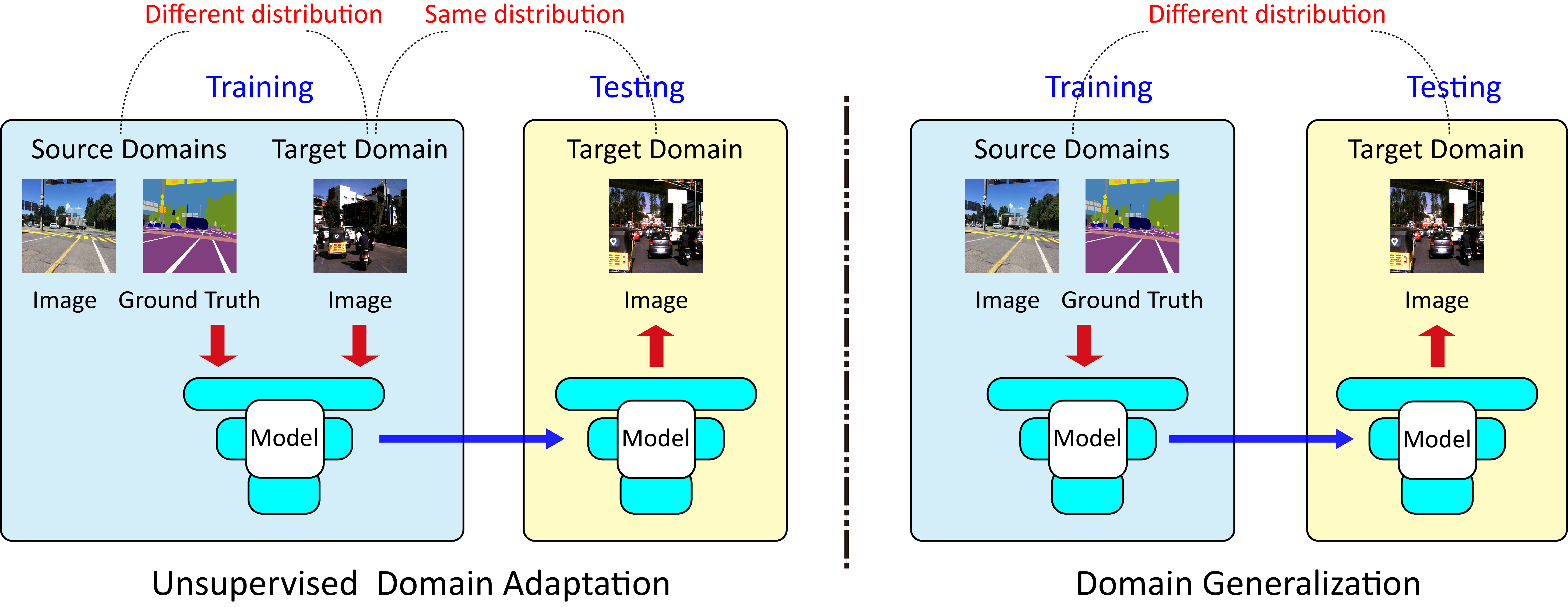} 
  \caption{Comparison between UDA and DG.} 
  \label{fig:introduction}
  \end{center}
\end{figure}

Notably, we expect that one model can be trained only once yet could generalize well in other unseen (or new coming) scenarios without any additional retraining procedure. With this expectation, deployment of the trained model to unseen scenarios becomes feasible, since the cost for data collection, data annotation, and model training can be simultaneously reduced. This setting is referred to as \textit{domain generalization} (DG) that considers how to acquire knowledge from an arbitrary number of related domains, and then apply the trained model directly to segment/classify previously unseen domains~\cite{muandet2013domain}. The key difference between UDA and DG setting relies on whether target domain data {could be observed and used} in the training stage, as shown in Fig.~\ref{fig:introduction}. Although DG has a similar goal with UDA, it is {much more challenging since we have no} available samples in the target domain. That is to say, the target domain in DG cannot be used, or even observed.

Recently, DG for image classification task has attracted lots of attention~\cite{yue2019domain,balaji2018metareg,li2019feature}. However, according to our best knowledge, only very few methods focus on the setting of DG in semantic segmentation (\aka, generalizable semantic segmentation). For example, IBN-Net~\cite{pan2018two} exploits IN to alleviate the style discrepancy across different domains. Also, the method in~\cite{yue2019domain} transfers the synthetic images to real images with auxiliary datasets and learns the domain-invariant representations with augmented dataset effectively. 
We notice that these methods all utilize the plain training paradigm, which is easy to overfit the source domains. Besides, during the test stage, the statistics of the normalization are from the source domain, which could suffer from the domain shift. 



Different from these aforementioned methods, we address domain generalization of semantic segmentation from 1) \textit{training paradigm} and 2) \textit{test strategy} perspectives together. We put forward a novel domain generalization framework to help model well generalize to the unseen target domain in semantic segmentation.

In particular, \textit{for the training paradigm}, we adopt the model-agnostic learning method (\ie, meta-learning for domain generalization (MLDG)~\cite{li2018learning}) to simulate the domain shift problem with episodic training paradigm, which has demonstrated its superior performance in image classification. It can be regarded as a regularization mechanism that prevents the model from overfitting the source domains.

\textit{For the test strategy}, we propose target-specific normalization to alleviate the data-distribution discrepancy between seen source and unseen target domains in the test stage. Specifically, regardless of training CNN in supervised, semi-supervised, unsupervised, or even the domain adaptation setting, we often use the mean and variance calculated from the training samples to conduct the batch normalization in the test stage. {Since the distribution of source domains is different from the unseen domains, normalizing the unseen domains with these statistics could suffer from the domain shift problem. To address this issue in the DG setting, we utilize the mean and variance calculated from the target domain during the test procedure to perform normalization, which can adapt to various unseen domains without any finetuning step on the network.}


However, the inaccurate estimation of statistics in batch normalization might degrade the performance, which is caused by the small batch size~(especially for the batch size of 1 {in the test stage of DG}). 
To boost the performance of the network with an accurate estimation, we propose to store previously tested images as an image-based memory bank~(abbreviated as Image Bank in the following part). When test images come one by one, we select images from the Image Bank for an accurate statistics estimation of the current sample with style-based selection policy. 

Extensive experiments demonstrate that the proposed method produces state-of-the-art performance for the domain generalization of semantic segmentation on multiple benchmark segmentation datasets, \eg, our method improves up to 5.54\% based on a strong baseline. Moreover, according to our ablation study, we sufficiently validate the effectiveness of each component in the proposed method.

Our contributions can be summarized as follows:
\begin{enumerate}
  \item We develop a novel domain generalization framework that jointly exploits the model-agnostic training scheme and the target-specific normalization test strategy to address the generalizable semantic segmentation task.
  \item {To deal with the inaccurate statistics estimation for a single sample in the test stage, we propose to utilize the Image Bank with the style-based selection policy to obtain accurate statistics.}
  \item We provide a strong baseline for semantic segmentation DG problem. Moreover, the proposed method achieves state-of-the-art performance for DG of semantic segmentation on multiple benchmark datasets. 
\end{enumerate}
\section{Related Work}


We review the previous literature from the related topics, \ie, unsupervised domain adaptation, domain generalization, and normalization. 

\textbf{Semantic segmentation.} Semantic segmentation belongs to a pixel-level classification task which is a fundamental yet challenge problem in computer vision and pattern recognition field. For accurate segmentation, several works aggregate multi-scale contextual information~\cite{peng2020semantic,zhang2021gpnet} and some of them refine low-resolution feature maps with high-resolution feature maps. Recently, more attention has been paid to the long-range dependency~\cite{li2020spatial,wang2021efnet} between pixels for better segmentation. Although these methods can achieve promising segmentation accuracy in supervised setting when training and test domains belong to the same domain, how to generalize well to other unseen domains in semantic segmentation is still an open problem.

\textbf{Unsupervised domain adaption.} Domain adaptation (DA) is a particular case of transfer learning. It leverages labeled data in one or more related source domains to learn a classifier for the target domain. Among these DA settings, the setting when the labeled data is not available in the target domain is referred to as unsupervised domain adaptation (UDA), which has been received considerable attention recently. Some methods first align distributions in pixel-level space \cite{chen2019crdoco}, and then adopt a generative network to narrow the distribution gap between source and target domains. Chang \etal \cite{chang2019all} aligns the feature distribution between source and target domains by adversarial training to keep semantic features consistent in different domains. Differently, Sun \etal \cite{sun2019not} weights the loss function by paying more attention to regions with similar label structures. However, since there is no available data in target domains, we cannot directly align the distribution of source and target domains in our generalizable semantic segmentation task. 
{Besides, some test time adaptation methods~\cite{sun2020test, wang2020fully} are proposed to deal with the domain shift problem in test time. During the test procedure, these methods and our method aim to better exploit the data distribution in the target domain, but there is an obvious diﬀerence that these methods all need to train the model with back-propagation in the test stage, while our method only does the inference without the cost of training model. Thus, the deployment of these methods is limited.}

\textbf{Domain generalization.} Different from UDA, domain generalization (DG) is a more challenging setting, which cannot use or even observe any target images during the training process. Current DG methods can be roughly classified into three categories: \emph{data augmentation}, \emph{domain invariant feature learning} and \emph{regularization}. 

The augmentation-based methods try to augment source domains with diverse styles. The model trained with these augmented data can be robust to the unseen domains with unseen styles. The techniques of augmentation are mainly based on Generative Adversarial Networks (GANs~\cite{goodfellow2014generative})~\cite{yue2019domain} or Adaptive Instance Normalization~(\eg, AdaIN~\cite{huang2017arbitrary})~\cite{somavarapu2020frustratingly}. One of the segmentation method~\cite{yue2019domain} in DG adopts CycleGAN to generate novel samples with image content from source domains and style from ImageNet.

The domain invariant feature learning-based methods try to learn an invariant representation across source domains. The representation should be able to generalize well to the unseen domain via the invariant hypothesis. All these methods try to first align the distribution by adversarial training~\cite{matsuura2020domain} or learning domain-shared and domain-specific components explicitly. Then they classify the unseen sample with only domain-shared component~\cite{d2018domain}.

Regularization-based methods assume the overfitting problem largely hurts the generalization ability. Thus they aim to reduce the overfitting problem during training via the regularization mechanism. Carlucci~\etal~\cite{carlucci2019domain} adopt self-supervised learning as a regularization term to help the generalization. Huang~\etal~\cite{huang2017arbitrary} propose to dropout the largest gradient during training to encourage distributed features and prevent overfitting. Meta-learning~\cite{finn2017model} based methods~\cite{balaji2018metareg,li2019feature,li2018learning} can also be seen as an approach of regularization due to its episodic training mechanism. It imitates the DG procedure where the model trained in seen domains should perform well in unseen domains. It has been applied to the DG classification task successfully.  For example, methods in~\cite{balaji2018metareg,li2019feature} are proposed to add a parameterized regularization term in meta-test stage to prevent the model from overfitting, while MLDG~\cite{li2018learning} directly adopts the MAML \cite{finn2017model} training framework. Although these DG methods performs well on the classification task, few works are developed to solve the generalizable semantic segmentation task. 


\textbf{Normalization.} Batch normalization (BN) \cite{ioffe2015batch} and instance normalization (IN) \cite{ulyanov2016instance} have been popularly used to stabilize training procedure and ease gradient explosion and vanishing problem. The literature~\cite{seo2019learning} shows that BN can preserve discriminative information in a domain, while IN is effective in reducing appearance difference and feature divergence~\cite{pan2018two}. Thus, recent approaches attempt to learn independent BN for each domain~\cite{seo2019learning,chang2019domain} and employ IN~\cite{pan2018two} to alleviate the domain gap. Besides, the method in~\cite{nam2018batch} leverages both BN and IN to exploit their advantages. However, all the above methods use the accumulated statistics (\ie, mean and variance) in source domains to normalize the images from target domains, which could suffer from the domain-shift problem in DG due to no available data in the target domain. {Recently, AdaBN~\cite{li2016revisiting} is proposed in UDA to utilize the statistics of the target domain to normalize images from the target domain. However, AdaBN calculates the statistics from all the test samples, which is unavailable in DG, while our method only requires a single image in the test stage.}

\begin{figure*}
  \centering    
  \includegraphics[scale=0.18]{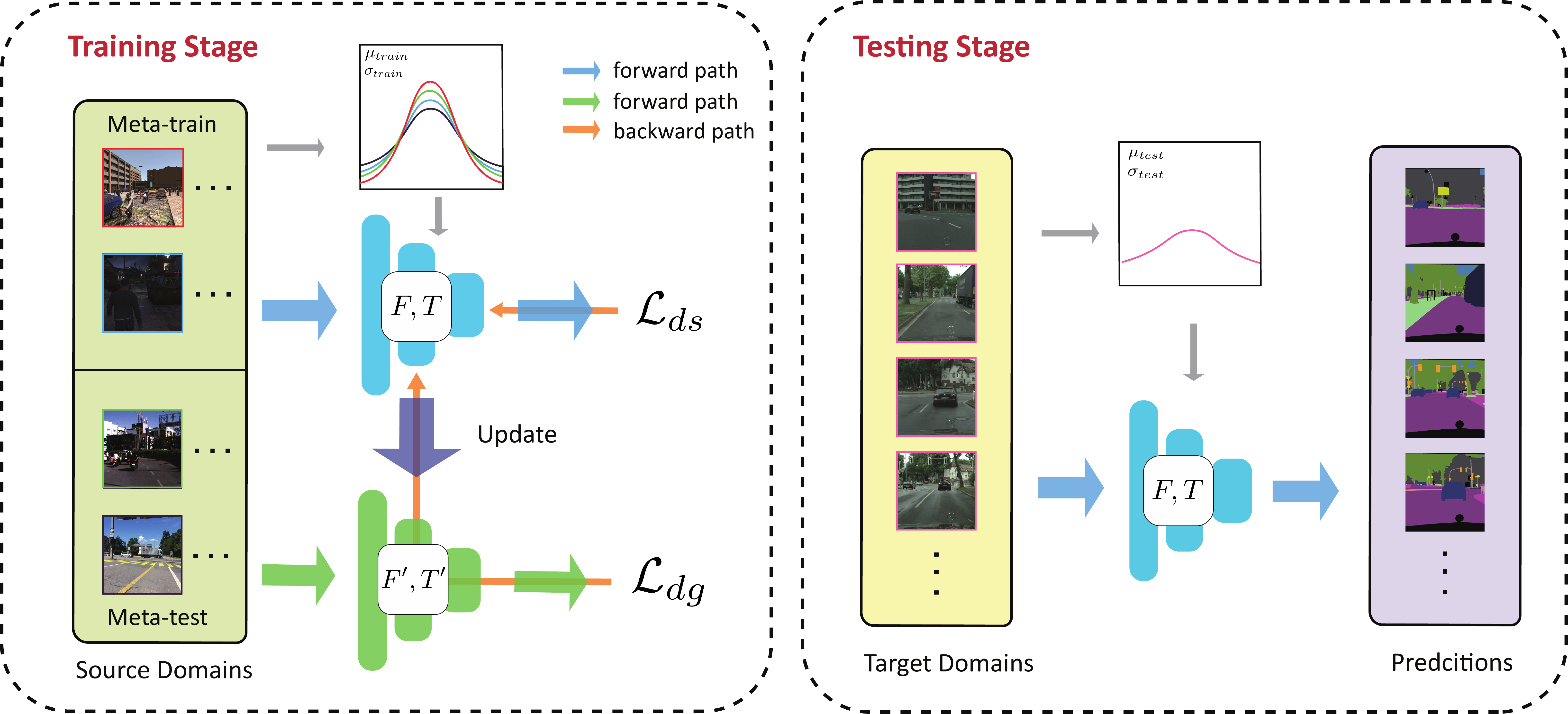} 
  \caption{The framework of our proposed method.} 
  \label{fig:structure}
\end{figure*}

\section{Our Method}
In this section, we introduce the technical details of our method which consists of 1) the model-agnostic learning scheme in the training stage, 2) the target-specific normalization and 3) the Image Bank in the test stage, as illustrated in Fig.~\ref{fig:structure}. In the following part, we first introduce notations in DG and present a simple method as a DG semantic segmentation baseline. We then formulate the meta-learning algorithm framework for semantic segmentation. {Afterwards, we describe the target-specific normalization in the test stage. Lastly, we detail the Image Bank in the test stage.}

\subsection{Our Baseline}
Let $\mathcal{X}$ denote the image space and $\mathcal{Y}$ denote the label space.
A domain is defined by a set of image and label pairs $\mathcal{D}^i = \{(x_j^i, y_j^i)\}_{j=0}^{N_i} $, where $x_j^i \in \mathcal{X}, y_j^i \in \mathcal{Y}$, and $N_i$ is the number of samples in the domain $\mathcal{D}^i$.
Assume that we have $N$ domains. $N-1$ domains are source domains $\mathcal{D}_{src} = \{\mathcal{D}^i\}^{N-1}_{i=1}$ and the left one domain is the target domain $\mathcal{D}_{target} = \{\mathcal{D}^N\}$. DG methods train the network with $\mathcal{D}_{src}$ and test in $\mathcal{D}_{target}$. 

Suppose we have a network to be trained. It contains a feature extractor $F(x;\theta)$ parametrized by $\theta$ and a classifier $T(f;\phi)$ parametrized by $\phi$. For image semantic segmentation, we usually use the cross-entropy (\emph{CE}) as loss function, which can be defined over dataset as follows:

\begin{equation}
  \label{eq01}
  \begin{aligned}
    \mathcal{L}(\mathcal{D}; \theta, \phi) = \frac{1}{|\mathcal{D}|}\sum_{(x, y)\in \mathcal{D}} \mathcal{L}_{CE}(T(F(x;\theta); \phi), y).
  \end{aligned}
\end{equation}



A straightforward DG method called AGG is introduced to train the model with $\mathcal{D}_{src}$ directly. It means simply aggregating all training data. We expect to minimize the following cross-entropy loss in Eq. (\ref{eq02}).

\begin{equation}\label{eq02}
\begin{aligned}
\mathcal{L}_{agg} = \mathcal{L}(\mathcal{D}_{src}; \theta, \phi).
\end{aligned}
\end{equation}

Note that AGG is a promising baseline in the semantic segmentation of domain generalization when $\mathcal{D}_{src}$ consists of multiple source domains. We will show the experimental results in Section \ref{experiments}.

\subsection{Model-agnostic Meta-learning in DG}

We can further improve the above baseline (\ie, AGG) by using the model-agnostic learning scheme. Concretely, we adopt MLDG (\ie, meta-learning for domain generalization)~\cite{li2018learning} to train our model. MLDG follows the MAML framework and has been successfully utilized in the DG classification task. In MAML, we construct many tasks from the auxiliary dataset, and each task contains a support set and query set. We hope that the network fine-tuned with the support set can generalize well in the query set. Similar to MAML, the DG task expects that the network fine-tuned in seen domains can generalize well in unseen domains. Thus, at each iteration, we randomly sample a task consisting of a sample set $\mathcal{B}$ from each source domain. We partition $\mathcal{B}$ into the meta-train data $\mathcal{B}_{tr}$ and meta-test data $\mathcal{B}_{te}$, where $ \mathcal{B}_{tr} \cap \mathcal{B}_{te}=\emptyset, \mathcal{B}_{tr} \cup \mathcal{B}_{te}=\mathcal{B}$. These two sets have no intersection in domains and correspond to the support set and query set, respectively. We first calculate the domain-specific loss $\mathcal{L}_{ds}$ by 

\begin{equation}\label{eq03}
\begin{aligned}
\mathcal{L}_{ds} = \mathcal{L}(\mathcal{B}_{tr}; \theta, \phi).
\end{aligned}
\end{equation}
Basically, we can calculate the gradient $\triangledown \mathcal{L}_{ds} $ of $\mathcal{L}_{ds}$ with back-propagation. Then the new parameter $\theta'$ and $\phi'$ can be updated with SGD (\ie, stochastic gradient decent) as follows:
  \begin{equation}\label{eq04}
\begin{aligned}
  \theta' &= \theta -  \eta * \triangledown_{\theta} \mathcal{L}_{ds};\\
  \phi' &= \phi -  \eta * \triangledown_{\phi} \mathcal{L}_{ds},
\end{aligned}
\end{equation}
where $\eta$ is the inner learning rate during model updating. Similar to MAML, we expect the updated network can generalize well in unseen domain. In this way, the domain generalization loss $\mathcal{L}_{dg}$ is calculated with meta-test data $\mathcal{D}_{te}$ as follows:
\begin{equation}\label{eq05}
\begin{aligned}
  \mathcal{L}_{dg} = \mathcal{L}(\mathcal{B}_{te}; \theta', \phi').
\end{aligned}
\end{equation}
We use \textbf{both these two losses} to update the origin parameters $\theta$ and $\phi$. In this meaning, the network is expected to perform well in both seen and unseen domains. Formally, the final meta-learning loss is
\begin{equation}\label{eq06}
\begin{aligned}
\mathcal{L}_{meta} = \mathcal{L}_{dg} + \alpha \mathcal{L}_{ds},
  \end{aligned}
\end{equation}
where $\alpha$ is a weight parameter to balance these two terms. The whole network is updated with $\mathcal{L}_{meta}$ and outer learning rate $\gamma$ as described in Algorithm~\ref{alg01}.

\begin{algorithm}
  \caption{Model-agnostic learning for generalizable semantic segmentation}\label{euclid}
  \textbf{Input} 
   source data $\mathcal{D}_{src}$, network parametrized by $\theta, \phi$, hyperparameters $\eta, \alpha,\gamma$\\
  \textbf{Output} the trained network
  \begin{algorithmic}[1]\label{alg01}
    \STATE \textbf{repeat} \\
    \STATE \hspace{\algorithmicindent} sample a mini-batch $\mathcal{B}$ (\ie, a task) from  $\mathcal{D}_{src}$

      \STATE \hspace{\algorithmicindent} random split $\mathcal{B}$ into meta-train $\mathcal{B}_{tr}$ and meta-test data $\mathcal{B}_{te}$ 
      \\\hspace{\algorithmicindent} $\mathcal{B}_{tr} \cap \mathcal{B}_{te}=\emptyset, \mathcal{B}_{tr} \cup \mathcal{B}_{te}=\mathcal{B}_{src}$ 

      \STATE \hspace{\algorithmicindent} compute domain-specific loss: $\mathcal{L}_{ds}=\mathcal{L}(\mathcal{B}_{tr}; \theta, \phi)$
      \STATE \hspace{\algorithmicindent} $\theta' = \theta -  \eta * \triangledown_{\theta} \mathcal{L}_{ds}$
      \\\hspace{\algorithmicindent} $\phi' = \phi -  \eta * \triangledown_{\phi} \mathcal{L}_{ds}$
      \STATE \hspace{\algorithmicindent} compute domain-generalization loss: $\mathcal{L}_{dg}=\mathcal{L}(\mathcal{B}_{te}; \theta', \phi')$
      \STATE \hspace{\algorithmicindent} compute the overall loss: $\mathcal{L}_{meta} = \mathcal{L}_{ds}+\alpha\mathcal{L}_{dg}$

      \STATE \hspace{\algorithmicindent} $\theta = \theta -  \gamma * \triangledown_{\theta}\mathcal{L}_{meta} $
      \\ \hspace{\algorithmicindent} $\phi = \phi -  \gamma * \triangledown_{\phi} \mathcal{L}_{meta}$
    \STATE \textbf{until} converge
  \end{algorithmic}
\end{algorithm}

\subsection{Target-specific Normalization}

In the conventional supervised, semi-supervised, unsupervised and domain adaptation scenarios, we normally use the accumulated mean $\hat{\mu} \in \mathbb{R}^{C}$ and variance $\hat{\sigma}^2  \in \mathbb{R}^{C}$ in the training stage to conduct normalization for the test data $x \in \mathbb{R}^{N\times C \times H \times W}$ as follows:

\begin{equation}\label{eq9}
\begin{aligned}
  \hat{x}_{n,c,h,w} = \frac{x_{n,c,h,w} - \hat{\mu}_c}{\sqrt{\hat{\sigma}_c^2 + \epsilon}}w_{c} + b_c.
\end{aligned}
\end{equation}
$N, C, H, W$ are number of images,  number of channels, image height and image width, respectively. $x_{n,c,h,w}$ denotes the $n\times c\times w\times h$-th element. 
However, in our DG setting, the statistics in source domains is different from target domain due to the domain gap. Considering this fact, we propose the target-specific normalization (TN) to directly use the statistics of target domain to normalize the features. Concretely, we obtain the new mean and variance for a mini-batch with $M$ samples in the test stage as follows:

 \begin{equation}\label{eq11}
\begin{aligned}
\bar{\mu}_c &= \frac{1}{MHW}\sum_{n=1}^{M}\sum_{h=1}^{H}\sum_{w=1}^{W} x_{n,c,w,h};\\
\bar{\sigma}_c^2 &= \frac{1}{MHW}\sum_{n=1}^{M}\sum_{h=1}^{H}\sum_{w=1}^{W} (x_{n,c,w,h}-\bar{\mu}_c)^2.\\
\end{aligned}
  \end{equation}

  Therefore, in the test stage, Eq. (\ref{eq10}) can be rewritten as below:
  \begin{equation}\label{eq10}
\begin{aligned}
  \hat{x}_{n,c,h,w} = \frac{x_{n,c,h,w} - \bar{\mu}_c}{\sqrt{\bar{\sigma}_c^2 + \epsilon}}w_{c} + b_c.
\end{aligned}
\end{equation}

\emph{Remark.} For the target-specific normalization (TN), we present some analysis and observations as follows: 1) In our experiments, we find that if we merely use the target-specific normalization on a model trained by the standard training method (\ie, non-MLDG training scheme), TN cannot effectively improve the generalization ability of the model in the unseen target domain; 
2) The size of a mini-batch has an impact on the final performance.  Besides, if $M>1$, the different sample combinations in a mini-batch might obtain different results. To address this issue, each experiment is repeated $5$ times, and the average results are reported.

\begin{figure}
  \centering  
  \begin{center}
  \includegraphics[scale=0.22]{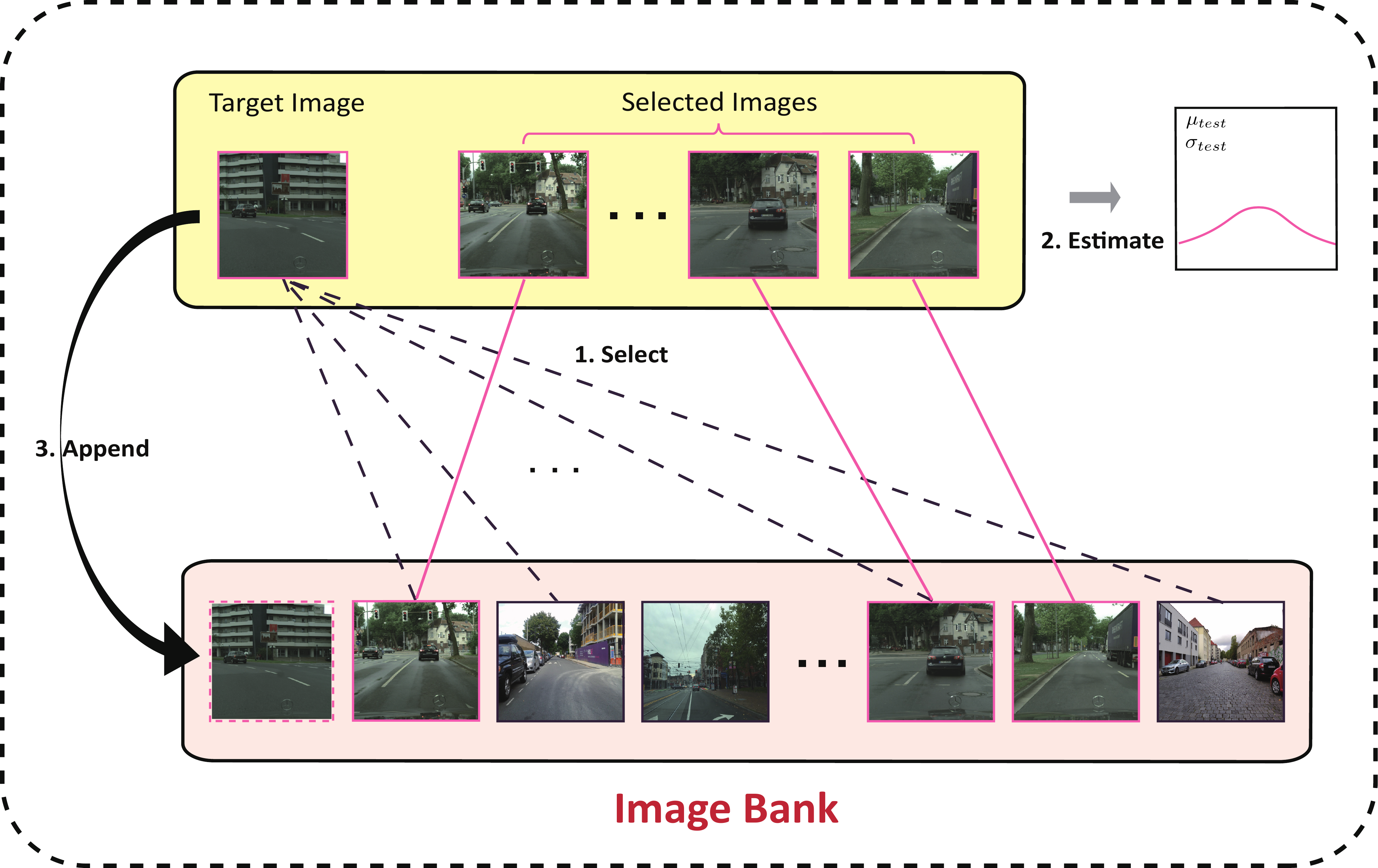} 
  \caption{Illustration of Style-based Selection policy for Image Bank.} 
  \label{fig:ib}
  \end{center}
  \vspace{-0.2cm}
\end{figure}

\subsection{Image Bank}
In real-world applications, several test images could come one by one. {We now provide an example about online image recognition for illustration: to predict the label for a given query image, if we deploy our trained model on a website, the user can merely upload one image to the website. In this situation, we have to process the uploaded images one by one for each user (\ie, batch size is 1 in test stage).} Although the target-specific normalization can adapt the model to the unseen domain with statistics calculated in the test stage, we argue that it cannot produce accurate statistics estimation for \textit{a single image} in the test stage. To obtain more accurate statistics, we put previously tested images into an Image Bank to aid the estimation of new coming images. Specifically, we adopt Image Bank with a fixed size of $Q$ to store these images. When an unseen image comes, we combine it with several previously tested images to produce its statistics. 
We now introduce our style-based selection policy in our Image Bank.

{\textbf{Style-based selection policy.} We notice that simply utilizing the previously stored images in the Image Bank could suffer from the domain shift problem when images in the Image Bank are from different domains. It might also result in an inaccurate estimation. The images from similar domains have similar style~\cite{yue2019domain}. and data distribution. And they can help robust statistics estimation. Being aware of this fact, we propose the Style-based selection policy for Image Bank~(SIB) to select the images that are similar to the current image in the style, as illustrated in Fig.~\ref{fig:ib}. Specifically, as aforementioned, since the normalization statistics represent the style~(distribution) information~\cite{huang2017arbitrary,nam2018batch}, we adopt $s=(\mu, \sigma)$ in the statistics to represent the style information in a given image. Considering the lower layers capture more detailed style information, to better represent the style information, we first extract the style statistics from the first layer of ResNet. Then, we calculate the symmetric KL Divergence~\cite{pan2018two} between styles of the current image and the images in the Image Bank as follows:
\begin{equation}
  \begin{aligned}
    D(s_i || s_j)  &= KL(s_i||s_j) + KL(s_j||s_i); \\
    KL(s_i || s_j) &= \log{\frac{\sigma_j}{\sigma_i}} + \frac{\sigma_i^2+(\mu_i-\mu_j)^2}{2\sigma^2_j}-\frac{1}{2}.
  \end{aligned}
\end{equation}
Finally, we select the top $M$ similar images from Image Bank and concatenate them with the current image to generate the statistics for the current image.} {When the Image Bank is full, we simply remove the out-of-date images with First In, First Out (FIFO) strategy. We also employ the random removal strategy and find no obvious difference in performance. Since the order in which the test images arrive is random, FIFO could equal to random removal scheme.}

\section{Experiments}\label{experiments}
In this section, we report both the quantitative and qualitative results. Specifically, we first describe the detail of datasets and implementation. Then, we extensively compare our method with state-of-the-art DG and UDA methods. Moreover, we conduct an ablation study to confirm the effectiveness of each module used in our framework. Lastly, we analyze the properties of our method systematically.

\subsection{Dataset and Implementation Details}

\textbf{Dataset.} We totally introduced five semantic segmentation datasets including: Cityscapes~\cite{cordts2016cityscapes}, GTA5~\cite{richter2016playing}, Synthia~\cite{ros2016synthia}, IDD~\cite{varma2019idd} and Mapillary~\cite{neuhold2017mapillary} for the evaluation. 
Note that all these five datasets have the same class space (\ie, they have the same 19 classes). For the simplicity, the five domains are denoted as \textbf{G}, \textbf{S}, \textbf{I}, \textbf{M}, \textbf{C} for GTA5, Synthia, IDD, Mapillary and Cityscapes, respectively, in the following parts. The detail of each dataset is as follows:

\begin{itemize}
    \item \textbf{Cityscapes (C)} is a large-scale real-world street scene dataset consisting of 2,975 training, 500 validation and 1,525 test images, respectively. All the images are high resolution with a size of 2048 $\times$ 1024.
    \item \textbf{GTA5 (G)} is collected from the GTA5 game, resulting in accurate pixel-wise semantic labels. There are 24,966 images with a size of 1914 $\times$ 1052. 
    \item \textbf{Synthia (S)} comprises synthesized images that are generated by rendering a virtual city. We use a subset of this dataset called SYNTHIA-RAND-CITYSCAPES, including 9,400 images with the resolution of 1280 $\times$ 760.
    \item \textbf{IDD (I)} is more diverse compared with cityscapes. A total of 10,004 images are captured from Indian roads with the resolution of 1678$\times$968.
    \item \textbf{Mapillary (M)} is regarded as the largest dataset with 25,000 high-resolution images. The images are collected from all over the world and taken from a diverse source of image capturing devices. The dataset contains 18,000 training images, 2,000 validation images, and 5,000 test images, respectively. The size of all these images is larger than 1920 $\times$ 2080.
\end{itemize}

\noindent\textbf{Implementation details.}
 We use ResNet-50~\cite{he2015deep} pretrained on ImageNet~\cite{deng2009imagenet} in this work. Following the conventional semantic segmentation setting, we use a dilated version of ResNet, resulting in an output stride of 8. 
 The inner and outer learning rates of $\eta, \gamma$ are set to 1e-3 and 5e-3, respectively. The weight $\alpha$ is set to 1. 
 We adopt the SGD optimizer, where the weight decay and momentum are set to 0.9 and 5e-4, respectively. We employ the Poly scheduler to decrease the outer learning rate with a weight of $0.9$. 
 We adopt extensive online data augmentation methods, including random flip, random scales at the range of [0.5, 2.0], random Gaussian blur, and random crop with a size of $600 \times 600$. We use mean IoU as our evaluation metric. 
 The total training epoch is 120. At each iteration, the images are randomly sampled from each domain, forming the image size of $B\times D \times C \times H \times W$, where $B$, $D$, $C$, $H$, and $W$ are batch size, the number of domains, channels, height, and width, respectively. We use a batch size of 8 for training. {The test batch size $M$ and queue size $Q$ are set to 4 and 128 respectively.} When applying target-specific normalization, due to the variance of statistics in a mini-batch, all results are averaged in five runs with different random seeds in the test stage. We evaluate our method in the validation set if not mentioned. 
 
\subsection{Comparison with the States-of-the-arts}

\begin{table}[t]
  \renewcommand\arraystretch{1}
  \scriptsize
     \caption{Comparison with state-of-the-arts on Cityscapes. Note that ``AGG-4'' and ``AGG'' represent the baseline in the ``G+S+I+M $\rightarrow$ C'' task and the ``G $\rightarrow$ C'' task. }
  \begin{center}
     \label{tab:state-of-the-art}
     \begin{tabular}{c|c|cc|c}
        \toprule
        ~~Methods~~  & ~~Backbone~~ & ~~mIoU~~ & ~~$\uparrow$~~  & ~~Publication~~\\
        \hline
        \multirow{2}{*}{}AGG & \multirow{2}{*}{ResNet-50} & 22.17 & \multirow{2}{*}{7.47} & \multirow{2}{*}{ECCV2018}\\
        IBN-Net~\cite{pan2018two}                  &                            & 29.64 &  \\
        \hline
        \multirow{2}{*}{}AGG & \multirow{2}{*}{ResNet-50} & 32.45 & \multirow{2}{*}{4.97} & \multirow{2}{*}{ICCV2019}\\
        Yue \emph{et al.}\cite{yue2019domain}    &             & 37.42 & \\
        \hline
        \multirow{2}{*}{}AGG & \multirow{2}{*}{ResNet-50} & 36.94 & \multirow{2}{*}{5.97} & \multirow{2}{*}{This paper}\\
        Our Method               &                            & 41.09 &  \\
        \hline
        \multirow{2}{*}{}AGG-4 & \multirow{2}{*}{ResNet-50} & 55.65 & \multirow{2}{*}{5.54} & \multirow{2}{*}{This paper}\\
        Our Method                &                            & \textbf{61.19} &  \\
        \bottomrule
     \end{tabular}
  \end{center}
\end{table}

We compare our method with two state-of-the-art DG segmentation methods~\cite{pan2018two,yue2019domain}. 
These two methods use GTA5 as the source domain and Cityscapes as the target domain. IBN-Net~\cite{pan2018two} plugs instance normalization into ResNet to remove style information. The method in~\cite{yue2019domain} augments GTA5 dataset with various styles transferred from ImageNet. Thus the performance in~\cite{yue2019domain} is better due to the larger dataset. For our method, we conduct experiments in two settings. The first one is using a similar setting to the literature~\cite{yue2019domain}. The second one is using GTA5, Synthia, Mapillary, and IDD as source domains, \ie, the ``G+S+I+M $\rightarrow$ C'' task. We evaluate our method in the test set of Cityscapes, as reported in Table~\ref{tab:state-of-the-art}. Note that ``AGG'' in this table denotes the baseline model. It uses the ResNet-50 as the backbone and is trained only in the source domain. ``AGG-4'' in Table~\ref{tab:state-of-the-art} indicates the baseline in the second setting. Firstly, in the same setting, our method can significantly outperform the method in~\cite{yue2019domain}. For example, based on the higher baseline (\ie, ($36.94$ vs. $32.45$)), our method can still obtain the more improvement (\ie, ($5.97$ vs. $4.97$)). Secondly, in the ``G+S+I+M $\rightarrow$ C'' task, we gain a strong baseline. Moreover, the result can be further risen by $5.54\%$ ($61.19$ vs. $55.65$) on Cityscapes, which shows the effectiveness of our method. {Note that our method also does not use the test data. When applying target-specific normalization, we only utilize \textit{previously seen test images} instead of the whole test data to estimate the statistics. Therefore, the comparison is fair.}


\subsection{Comparison with Unsupervised Domain Adaptation Methods} 

To further exhibit the superiority of our method, we also compare our method with some unsupervised domain adaptation methods. The results are reported in Table \ref{tab:uda}, which is obtained by submitting the predictions in the test set of Cityscapes. All these UDA methods are trained with labeled GTA5 or Synthia dataset and unlabeled Cityscapes dataset. To compare with them, due to the limitation of our methods that the number of source domains should bigger than 1, we train our model with both GTA5 and Synthia. As seen in Table \ref{tab:uda}, our method can achieve comparable performance to those UDA methods. {Note that our method does not employ any data from the target domain to train the model. This disadvantage further highlights the superiority of our method.} Also, our method is more valuable than the UDA methods in the real-world application, because we do not need to collect any data from a new scenario (\ie, the unseen target domain).

\begin{table*}[t]
  \renewcommand\arraystretch{0.8}
  \normalsize
  \caption{Comparison to unsupervised domain adaptation methods. The first group methods are in the ``GTA5 $\rightarrow$ Cityscapes'' task. The second group methods are in the ``Synthia $\rightarrow$ Cityscapes'' task.}
  \begin{center}
     \label{tab:uda}
     \scalebox{0.6}{
      \begin{tabular}{c|ccc|ccc|cc}
      \toprule
      Methods 
      & \rotatebox[origin=c]{90}{Tsai et al.\cite{tsai2018learning}} 
      & \rotatebox[origin=c]{90}{SAPNet\cite{li2020spatial}} 
      & \rotatebox[origin=c]{90}{AdvEnt\cite{vu2019advent}} 
      
      & \rotatebox[origin=c]{90}{Tsai et al.\cite{tsai2018learning}} 
      & \rotatebox[origin=c]{90}{AdvEnt\cite{vu2019advent}} 
      & \rotatebox[origin=c]{90}{SAPNet\cite{li2020spatial}} 
      
      & \rotatebox[origin=c]{90}{Our method} 
      & \rotatebox[origin=c]{90}{Our method} \\
          \hline 
          Base model & Deeplabv2& Deeplabv2 &Deeplabv2 &Deeplabv2 & Deeplabv2 & Deeplabv2 & ResNet-50 & Deeplabv2\\
          \hline 
          road      & 86.5 & 88.4 & 89.4 &    84.3 & 85.6 & 81.7    & 85.3 & 85.8 \\
          sidewalk  & 36.0 & 38.7 & 33.1 &    42.7 & 42.2 & 33.5    & 42.8 & 37.4 \\
          building  & 79.9 & 79.5 & 81.0 &    77.5 & 79.7 & 75.9    & 77.3 & 74.1 \\
          wall      & 23.4 & 29.4 & 26.6 &    9.3  & 8.7  &  7.0    & 19.2 & 15.1 \\
          fence     & 23.3 & 24.7 & 26.8 &    0.2  & 0.4  &  6.3    & 20.9 & 12.0 \\
          pole      & 23.9 & 27.3 & 27.2 &    22.9 & 25.9 & 74.8    & 32.2 & 37.8 \\
    traffic light   & 35.2 & 32.6 & 33.5 &    4.7  & 5.4  & 78.9    & 44.8 & 36.8 \\
    traffic sign    & 14.8 & 20.4 & 24.7 &    7.0  & 8.1  &  6.3    & 26.4 & 25.2 \\
    vegetation      & 83.4 & 82.2 & 83.9 &    77.9 & 80.4 & 74.8    & 86.0 & 84.2 \\
          terrain   & 33.3 & 32.9 & 36.7 &     -   &  -   &   -     & 46.9 & 47.5 \\
          sky       & 75.6 & 73.3 & 78.8 &    82.5 & 84.1 & 78.9    & 84.4 & 87.8 \\
          person    & 58.5 & 55.5 & 58.7 &    54.3 & 57.9 & 52.1    & 67.6 & 71.7 \\
          rider     & 27.6 & 26.9 & 30.5 &    21.0 & 23.8 & 21.3    & 18.6 & 30.7 \\
          car       & 73.7 & 82.4 & 84.8 &    72.3 & 73.3 & 75.7    & 85.4 & 81.8 \\
          truck     & 32.5 & 31.8 & 38.5 &      -  &   -  &  -      & 17.3 & 17.0 \\
          bus       & 35.4 & 41.8 & 44.5 &    32.3 & 36.4 & 30.6    & 18.1 & 25.2 \\
          train     &  3.9 &  2.4 &  1.7 &     -   &   -  &   -     & 0.6  & 3.64 \\
       motorcycle   & 30.1 & 26.5 & 31.6 &    18.9 & 14.2 & 10.8    & 26.4 & 34.2 \\
          bicycle   & 28.1 & 24.1 & 32.4 &    32.3 & 33.0 & 28.0    & 15.2 & 31.1 \\ 
          \hline 
          mIoU      & 42.4 & 43.2 & 45.5 &    40.0 & 41.2 & 44.3    & 42.9 & 44.2 \\
        \bottomrule
      \end{tabular}
     }
  \end{center}
\end{table*}

\begin{figure*} 
  \centering 
  \includegraphics[scale=0.17]{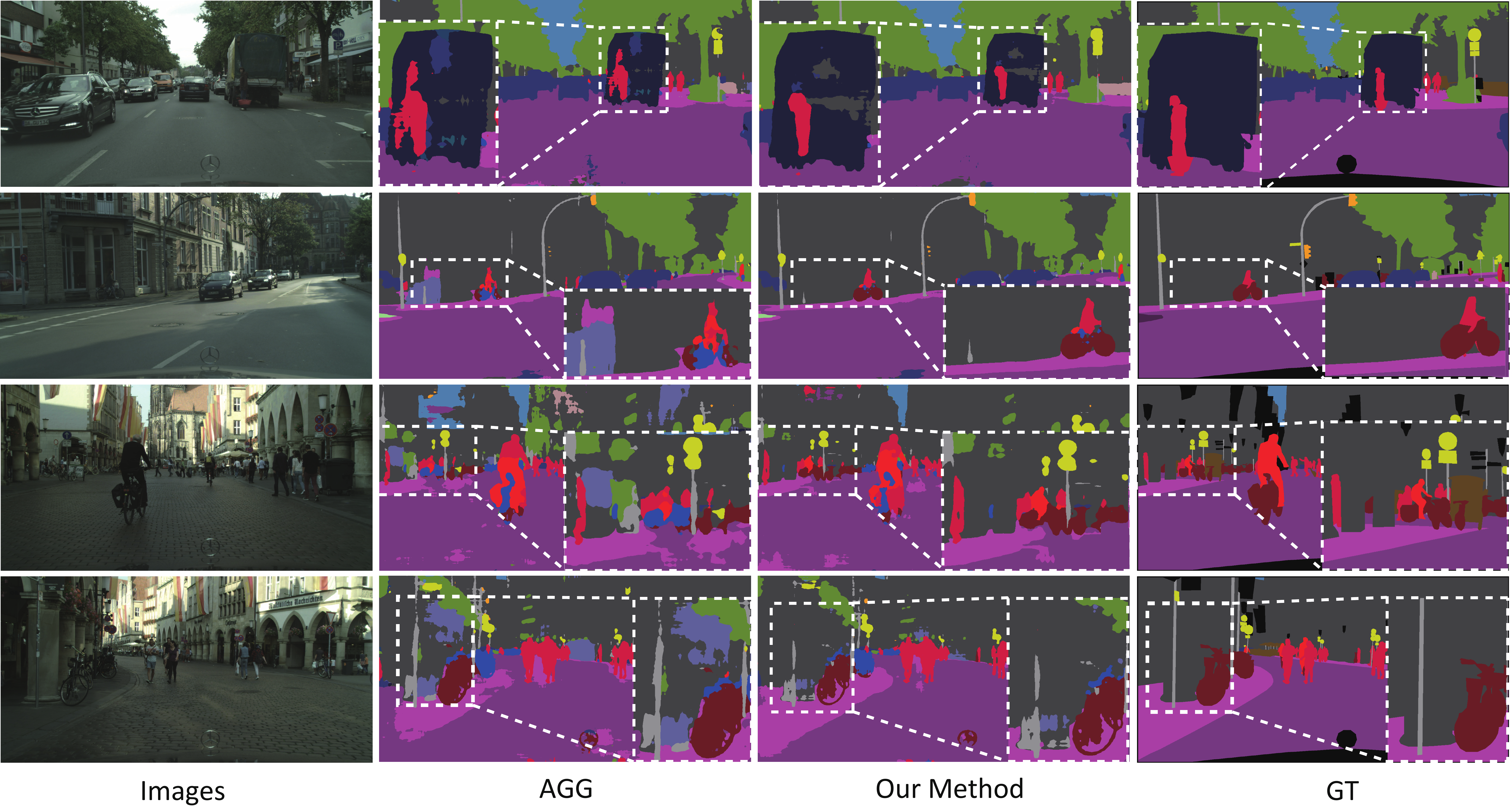}  
  \vspace{-0.8cm}
  \caption{Some visualization results from the validation set in Cityscapes.} 
  \label{fig:compare}
\end{figure*}

\subsection{Ablation Study}\label{sec:ablation}

To analyze the efficacy of each component, we perform ablation study on three and four source domains. We report the experimental results in Table~\ref{tab:AblationStudy}. Firstly, compared with ``AGG'', ``MLDG'' can obtain better performance, especially for the model trained on four source domains. For example, for the ``G+S+I+M $\rightarrow$ C''  task, we achieve improvement by $2.24\%$ ($55.95$ vs. $53.71$). Thus, this confirms the effectiveness of MLDG in DG semantic segmentation. Secondly, we observe that the target-specific normalization benefits from the model trained by the model-agnostic learning scheme. Specifically, in the ``G+S+I+M $\rightarrow$ C'' task, ``MLDG+TN'' improves ``MLDG'' by $1.24\%$ ($55.95$ vs. $57.19$), while ``AGG+TN'' has an inferior performance than ``AGG''. This comparison demonstrates that using the model-agnostic learning skill and target-specific normalization together can achieve good performance in the generalizable semantic segmentation task. {Finally, the results of ``MLDG+SIB'' outperform ``MLDG+TN'' in all tasks. It validates that the target-specific module with Image Bank can enhance the ability of generalization in the unseen target domain.} {Besides, Our method has stable results with small std (\eg, in the ``G+S+M+I$\rightarrow$C" task, the mean and std of our method are 57.66 $\pm$ 0.17), because the selection procedure in Image Bank tends to select most similar images, which can produce more stable statistics for each test image.} Some visual results from the model trained on four source domains are shown in Fig.\ref{fig:compare}. As seen in the white bounding box, the segmentation results of our method are closer to the ground truth when compared to the AGG.

\begin{table}[t]
  \renewcommand\arraystretch{1.1}
  \scriptsize
  
  \begin{center}
     \caption{Evaluation of the effectiveness of different modules in the proposed method. TN means adopt the target-normalization test and SIB means the Style-based selection policy for Image Bank.}
     \label{tab:AblationStudy}
     
     \resizebox{\columnwidth}{!}{
     \begin{tabular}{c|c|c|c|c}
     \toprule
      ~Methods~  &   ~G+S+I$\rightarrow$ M~  & ~G+S+I$\rightarrow$ C~& ~G+S+I+C $\rightarrow$ M~ & ~G+S+I+M $\rightarrow$ C~ \\
      \hline
      AGG      &  49.26  & 48.31 & 52.89 & 53.71 \\
      AGG+TN  & 41.24 & 46.21 & 41.75 & 51.90  \\
      AGG+SIB & 44.54 $\pm$ 0.22 & 48.17 $\pm$ 0.13 & 46.90 $\pm$ 0.34 & 54.48 $\pm$ 0.11\\

      \hline
      MLDG     &  49.00  & 48.64 & 53.85 & 55.95 \\
      MLDG+TN  & 50.11 & 50.72 & 53.32 & 57.19 \\ 
      MLDG+SIB~(Our Method) & \textbf{50.51 $\pm$ 0.20} & \textbf{51.49 $\pm$ 0.04} & \textbf{55.13 $\pm$ 0.19} & \textbf{57.66 $\pm$ 0.17}\\
      \bottomrule
     \end{tabular}
     }
  \end{center}
  
\end{table}

\subsection{Further Analysis}

In this paragraph, we conduct more experiments to further analyze the property of the proposed method from multiple different views. 

\textbf{Influence of the divided ratio of meta-train and meta-test.} In MLDG, we need to split source domains into meta-train and meta-test domains, \eg, we can split four source domains into three meta-train domains and one meta-test domain or one meta-train domain and three meta-test domains. To investigate the best option to split the four source domains, we enumerate the partition ratio of meta-train and meta-test, including ``3:1'', ``2:2'', and ``1:3''. The performance is shown in Table \ref{tab:split}. As seen, the best option is to split the source domains into ``2:2''. We hypothesize that the balanced partition scheme guarantees that both of the two models trained on mete-train and meta-test domains can achieve relatively good performance, thus the final model can better generalize in unseen domains. For all the experiments with four source domains in this paper, we set meta-train and meta-test to 2 and 2. Besides, the meta-test procedure is critically important when comparing the results of the ``3:1'' to ``1:3'' (\ie, using three source domains as meta-test produces better outcome), which imitates the process of domain generalization in the training stage. This meta-test step is also mainly different from the general training method that the model is directly trained using all source domains (\ie, the baseline, AGG).

\begin{table}[t]
  \renewcommand\arraystretch{1.2}
  \caption{Performance with the various divided ratios of meta-train and meta-test in the ``G+S+I+M $\rightarrow$ C'' task. $n:m$ means source domains are split into $n$ meta-train domains and $m$ meta-test domains.}
  \scriptsize
  \begin{center}
     \label{tab:split}
     \resizebox{0.6\columnwidth}{!}{
       \begin{tabular}{c|ccc}
       \toprule
        ~~~~$n:m$~~~~  & ~~~~3:1~~~~    & ~~~~2:2~~~~   & ~~~~1:3~~~~ \\ 
        \hline
        mIoU     & 54.38  & \textbf{55.95} & 55.41 \\
        \bottomrule
     \end{tabular}
     }
  \end{center} 
\end{table} 

\begin{figure*}
  \centering
  \includegraphics[scale=0.45]{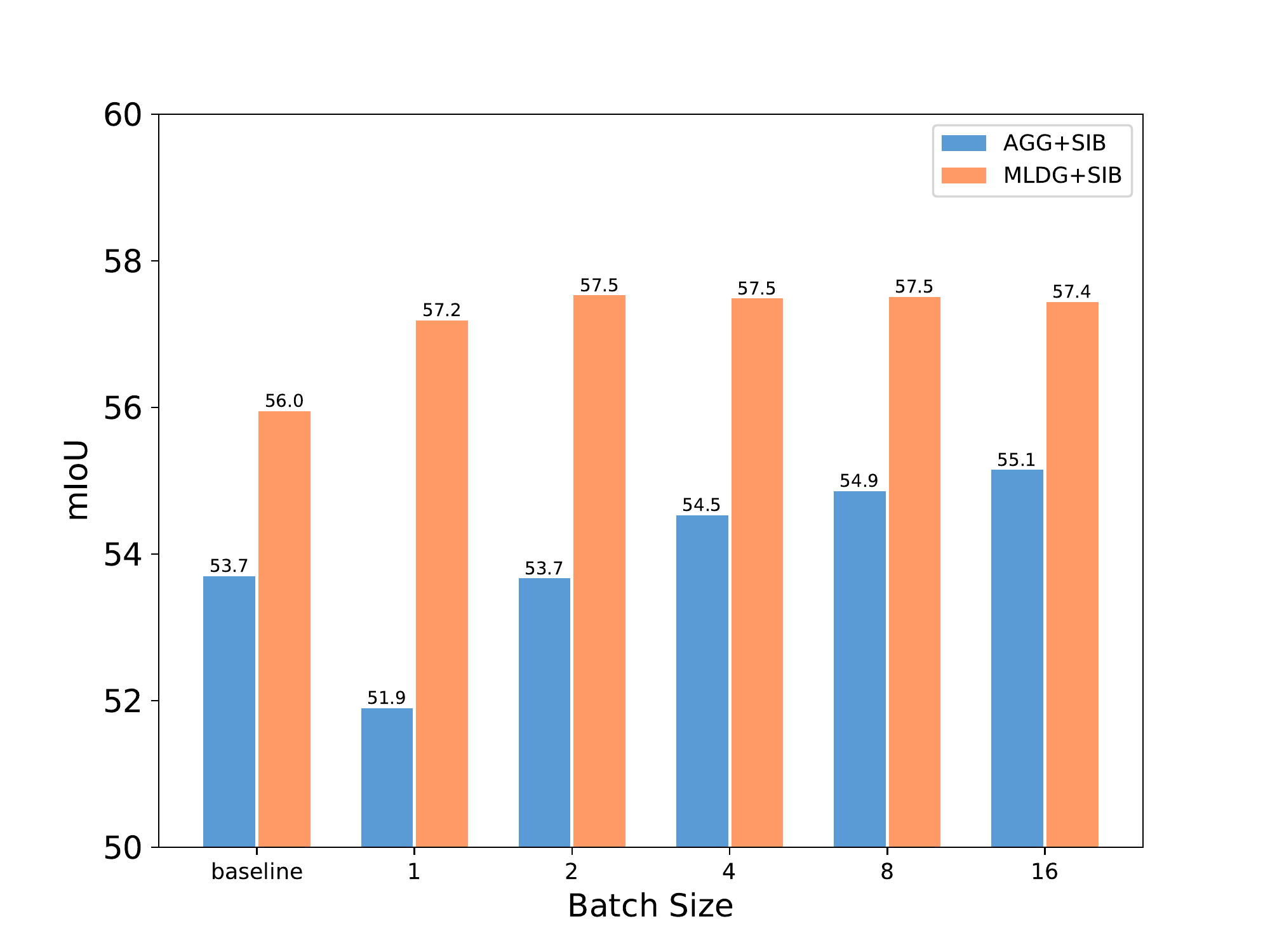} 
  
  \caption{The results with the various batch sizes in the ``G+S+I+M $\rightarrow$ C'' task. Note that the baseline in this figure denotes the model tested without target-specific normalization.} 
  \label{fig:batch-size}
\end{figure*}

\textbf{Evaluation of the batch size in target-specific normalization.} \label{sec:batch-size} In SIB, the batch size $M$ has an impact on the mean and variance. Thus we utilize different batch sizes to conduct the experiment and analyze the influence of different batch sizes in the test stage. The experimental results are illustrated in Fig.~\ref{fig:batch-size}.
As seen in this figure, with the increasing batch size, the performance gradually improves on the two models trained by AGG and MLDG. 
Note that An obvious trade-off is shown. When the batch size is increasing, the performance first increases and achieves the best performance, then decreases a little afterward. We hypothesize that there is a balance between specificity and commonality which corresponds to the image-specific style and domain-specific style in an image. When the batch size is 1, the statistics are solely based on the single image which is highly related to the image-specific style. When the batch size is large enough, the statistics are averaged over many images, which removes the specificity of the current image and is most related to the domain-specific style. 


Besides, as seen from Fig.~\ref{fig:batch-size}, although target-specific statistics increases the performance of the AGG model of batch size from 1 to 16, the result drops drastically when compared with the baseline of our method which uses the statistics from the training stage (\ie, the conventional batch normalization). Furthermore, its improvement is also little when the batch size is small. Differently, when we apply target-specific statistics on the MLDG model, even the batch size of 1 can still be better than the baseline. This further demonstrates that MLDG can better show the superiority of the target-specific normalization in the generalizable semantic segmentation task. Therefore, leveraging them together is excellently critical in the proposed method. 

\textbf{The effectiveness of style-based selection policy.} When images come one by one, to better address the discrepancy of image distribution problem in the Image Bank, the Style-based selection policy for Image Bank~(SIB) is proposed to estimate the statistics with images from similar domains in the Image Bank. We evaluate its performance on ``G+S+I $\rightarrow$ C+M" task. The simple Queue-based Image Bank is denoted as QIB. As shown in Table \ref{tab:ssp}, when we use QIB, the performance drops comparing to ``MLDG+TN'' due to the inaccurate statistics estimated with images from multiple domains in the Image Bank. After we apply SIB, images from the same domain are selected and the accurate statistics produce better performance. In particular, if test images come from the same target domain, the image bank based method equals to the target-special normalization with multiple images in a batch, as shown in Fig.~\ref{fig:batch-size}.

\begin{table}[t]
  \renewcommand\arraystretch{1.2}
  
  \caption{The effectiveness of the style-based selection policy for Image Bank~(SIB) when images are from multiple domains in the Image Bank. The performance is evaluated on the ``G+S+I $\rightarrow$ C+M" task. The Queue-based selection policy for Image Bank is denoted as QIB.}
  \label{tab:ssp}
  \begin{center}
       \resizebox{0.7\columnwidth}{!}{
     \begin{tabular}{c|c|c|c|c}
     \toprule
      Methods & MLDG + BN & MLDG + TN & MLDG + QIB & MLDG + SIB\\
      \hline
      ~G+S+I$\rightarrow$C+M~  & 50.87 & 52.04 & 51.87 & \textbf{52.80 $\pm$ 0.14} \\
      \bottomrule
     \end{tabular}
     }
  \end{center}
\end{table}

\begin{table}[t]
  \renewcommand\arraystretch{1.2}
  \scriptsize
    \caption{Influence of different layers to capture the style information. The network is evaluated on the ``G+S+I$\rightarrow$C+M" task with shuffled datasets. Target-specific Normalization~(TN) and Style-based nselection policy for Image Bank~(SIB) are both employed. ``Layer-x" means we adopt x-th layer as the style representation layer for the similarity calculation in the SIB. The mean IoU and the selection accuracy are both listed. }
  \begin{center}
     \label{tab:layers}
       \resizebox{0.7\columnwidth}{!}{
     \begin{tabular}{c|c|c|c|c}
     \toprule
      ~Layers~& ~Layer-1~ & ~Layer-2~ & ~Layer-3~ & ~Layer-4~  \\
      \hline
      mIoU & \textbf{52.78} & 52.44 & 52.24 & 52.30 \\
      Selection Acc & \textbf{96.78} & 93.52 & 89.51 & 87.51 \\
      \bottomrule
     \end{tabular}
     }
  \end{center}
\end{table}

\textbf{Influence of different layers to capture the style information.} 
Although SIB can effectively deal with the inaccurate estimation of statistics of multi-domain images. The quality of the similarity in SIB is highly dependent on using which layer in Resnet to obtain the style information. We conduct an experiment to investigate its influence. The mean IoU and the accuracy of selecting images from the same domain are all reported in Table~\ref{tab:layers}. It is observed that the best accuracy and segmentation performance are achieved when we choose the first layer to obtain the style information. Since it contains detailed style information, it is more accurate to calculate the similarity of styles between images.

\begin{table}[t]
  \renewcommand\arraystretch{1.2}
  
  \scriptsize
  \caption{The results of different queue sizes on the ``G+S+I$\rightarrow$C+M" tasks. Target-specific Normalization~(TN) and Style-based selection policy for Image Bank~(SIB) are both employed.}
  
  \begin{center}
    \label{tab:queue-size}     
     \begin{tabular}{c|c|c|c|c|c|c|c|c}
     \toprule
       Queue size & 512 & 256 & 128 &  64 &    32 &    16 &     8 &    4 \\
     \hline
      mIoU & 52.64 & 52.68 & \textbf{52.78} & 52.70 & 52.66 & 52.43 & 52.45 & 52.19 \\
      \bottomrule
     \end{tabular}

  \end{center}
\end{table}

  \textbf{Evaluation of queue size in Style-based Image Bank.} Besides that layer choice can affect the performance of SIB, the queue size also has an influence. As shown in Table~\ref{tab:queue-size}, with the queue size increases, the performance first increases and then decreases. We hypothesize that with a larger queue size, more images from the same domain can be selected for an accurate estimation of statistics. However, when selected images are too similar, the estimation will also be inaccurate because of little diversity in these images. Thus a relatively large queue size~(\eg, 128) can strike a balance.

\textbf{Influence on the number of source domains.} The number of source domains has a significant impact on the performance of DG. We conduct some experiments using a different number of source domains and utilize Cityscapes as the target domain. The results are reported in Table~\ref{tab:domain-number}. As seen, more source domains can achieve better performance. Besides, we observe that if the source domains have a similar domain with the target domain, the DG method can obtain significant improvement. For example, images from both GTA5 (G) and Synthia (S) are synthetic, and the images from IDD (I) and Mapillary (M) are collected from the real scenario, which are closer to the real dataset Cityscapes (C). Thus, the outcome from ``G+S $\rightarrow$ C'' to ``G+S+I $\rightarrow$ C'' has a larger advance than the performance from ``G $\rightarrow$ C'' to ``G+S $\rightarrow$ C''.

\begin{table}
  \renewcommand\arraystretch{1.2}
  \scriptsize
  \caption{Experimental results with different number of source domains. Note that Cityscapes is utilized as target domain.}
  \begin{center}
     \label{tab:domain-number}
     \begin{tabular}{c|cccc}
     \toprule
      ~~Source domain~~  & ~~G~~  & ~~G+S~~ & ~~G+S+I~~ & ~~G+S+I+M~~\\
      \hline
      ~~mIoU~~  & ~~29.33~~ & 38.10 $\pm$ 0.16  & 51.49$\pm$0.04 & 57.66$\pm$0.17\\
      \bottomrule
     \end{tabular}
  \end{center}
    \vspace*{-10pt}%
\end{table}

\begin{table}
  \renewcommand\arraystretch{1.2}
  \scriptsize
  \caption{Performance in source and target domains of the ``G+S+I+M $\rightarrow$ C'' task. SIB means the Style-based selection policy for Image Bank}
  \begin{center}
     \label{tab:source-domain}
     \scalebox{0.9}{
     \begin{tabular}{c|c|cccc}
      \toprule
      \multirow{2}[1]{*}{~~Methods~~}  & ~~Target~~      & \multicolumn{4}{c}{Source}\\
      \cmidrule{2-6}
                                    & ~~Cityscapes~~      & ~~IDD~~ & ~~Mapillary~~ & ~~Synthia~~ & ~~GTA5\\

      \hline
       AGG     & 53.71     & 60.42             & 61.25             & 59.49             & 56.73 \\
      AGG+SIB  &54.48 $\pm$ 0.11    & 55.07 $\pm$ 0.17 $\downarrow$ & 55.07$\pm$0.17   $\downarrow$ & 50.41 $\pm$ 0.11 $\downarrow$ & 48.00 $\pm$ 0.19 $\downarrow$ \\
      \hline
      MLDG       & 55.95           & 59.98          & 60.21           & 62.02          & \textbf{57.09} \\
      MLDG+SIB &  \textbf{57.66 $\pm$ 0.17} & \textbf{63.29 $\pm$ 0.14} & \textbf{63.81 $\pm$ 0.23}  & \textbf{63.99 $\pm$ 1.2} & 55.52 $\pm$ 0.07 \\
       \bottomrule
     \end{tabular}
     }
  \end{center}
\end{table}

\textbf{Analysis of meta-learning and target-specific normalization.} To further give prominence to the property of our method, we analyze the experimental results from the source domain perspective. We report the results in Table~\ref{tab:source-domain}. According to the comparison between the top group (``AGG'' and ``AGG+SIB'') and the bottom group (``MLDG'' and ``MLDG+SIB''~(Our Method)), the target-specific normalization cannot only enhance the generalizable ability in the unseen target domain but also improve the performance in source domains on the MLDG model. However, when we utilize SIB on the AGG model, the performance has a large deterioration in source domains. We suppose that the AGG model is sensitive to the statistics, while the model-agnostic scheme can guide the model to be more robust to it.

\renewcommand{\cmidrulesep}{0mm} 
\setlength{\aboverulesep}{0mm} 
\setlength{\belowrulesep}{0mm} 
\setlength{\abovetopsep}{0cm}  
\setlength{\belowbottomsep}{0cm}

\begin{table}
  \renewcommand\arraystretch{1.2}
  \scriptsize
  \caption{Experimental results with different backbones on the ``G+S+M+I$\rightarrow$C" task.}
  \begin{center}
     \label{tab:backbone}
     \begin{tabular}{c|cccc}
     \toprule
      ~~Source domain~~  & ~~Resnet50~~  & ~~DeepLabV3~\cite{chen2017rethinking}~~ & ~~HRNet~\cite{sun2019deep}~~\\
      \hline
      ~~MLDG~~         & 55.95   & 58.20      & 61.74\\
      ~~Our Method~~  & 57.66 $\pm$ 0.17 & 60.26 $\pm$ 0.10  & 64.09 $\pm$ 0.11 \\
      \bottomrule
     \end{tabular}
  \end{center}
    \vspace*{-10pt}%
\end{table}

\textbf{Effectiveness of our method in different backbones.} {To demonstrate that our method can easily be applied to different backbones, we have added the comparison to DeepLabV3~\cite{chen2017rethinking} and HRNet~\cite{sun2019deep}. From the Table \ref{tab:backbone}, we observe that using a stronger network can enhance the generalization ability of our method and our method can consistently improve based on other backbones (\eg, our method improves MLDG by 2.35\%~(64.09 vs. 61.74) based on HRNet). }

\begin{table}[t]
  \renewcommand\arraystretch{1.1}
  \scriptsize
  
  \begin{center}
     \caption{Comparison to AdaBN with our method on four different tasks. Our method and AdaBN are all performed based on MLDG for a fair comparison.}
     \label{tab:adabn}
     
     \resizebox{\columnwidth}{!}{
     \begin{tabular}{c|c|c|c|c}
     \toprule
      ~Methods~  &   ~G+S+I$\rightarrow$ M~  & ~G+S+I$\rightarrow$ C~& ~G+S+I+C $\rightarrow$ M~ & ~G+S+I+M $\rightarrow$ C~ \\
      \hline
      MLDG     &  49.00  & 48.64 & 53.85 & 55.95 \\
      AdaBN~\cite{li2016revisiting} & 50.29 & 50.31 & 54.10 & 56.67 \\
      Our Method & \textbf{50.51 $\pm$ 0.20} & \textbf{51.49 $\pm$ 0.04} & \textbf{55.13 $\pm$ 0.19} & \textbf{57.66 $\pm$ 0.17}\\
      \bottomrule
     \end{tabular}
     }
  \end{center}
  
\end{table}

\textbf{Comparison to AdaBN.} {We have compared our method to AdaBN~\cite{li2016revisiting} in Table \ref{tab:adabn}. Our method and AdaBN are all performed based on MLDG for a fair comparison. As reported from the table, our method can consistently outperform AdaBN on different tasks. As we have discussed in Section~\ref{sec:batch-size}, each image has its specificity and commonality in its corresponding domain. To normalize the image, there is a balance between specificity~(small batch size) and commonality~(large batch size). When the batch size is 1, the BN statistics estimation is solely based on the single image which cannot produce accurate statistics for its corresponding domain. When the batch size is large enough, the specificity in the statistics could be removed. Thus, our method can strike a balance between them and perform better than AdaBN. Particularly, AdaBN requires to collect all data from the target domain to do normalization, which is impractical in DG when only one image is provided at each moment in the test stage.}

\textbf{Inference time of the target-speciﬁc normalization.} {We report the time cost on Tesla V100 GPU with Resnet-50 as the backbone and input image size of 769 $\times$ 769. Since the semantic segmentation task is a time-consuming task due to the large output size, the normal test speed~(without SIB) is 0.12 seconds per image and when we apply SIB, the speed becomes 0.34 seconds per image. Although our method needs more time than the normal test, it can produce better performance~(\eg, in the ``G+S+M+I$\rightarrow$C" task, our method achieves 57.66 mIoU, which is better than 55.95 mIoU of the AGG method).}

{\textbf{Discussion:} In this paper, we consider two image processing scenarios including ``collect-collect-...-collect-process'' and ``collect-process-collect'' \textbf{\textit{in the test stage}}. When there are several test images available one-by-one (\ie, ``collect-process-collect'' scenario), the proposed Style-based selection policy of Image Bank (SIB) can calculate more accurate statistics and perform better than the target-specific normalization (TN) designed for handling the ``collect-collect-...-collect-process''  case, as shown in Table~\ref{tab:ssp}. Besides, compared with the typical batch normalization (BN), both our SIB and TN can achieve better performance in the ``one by one'' case.}

\section{Conclusion}
In this paper, we developed a novel domain generalization method to handle the generalizable semantic segmentation task, which jointly exploits the meta-learning training skill and the target-specific normalization from the model training and the data-distribution discrepancy perspectives. To deal with the inaccurate estimation of statistics in target-specific normalization, the Image Bank with the style-based selection policy was designed for an accurate estimation. By extensive evaluation, the efficacy of the proposed method was thoroughly validated from multiple different views. Besides, we got an interesting observation that the model trained by the model-agnostic learning skill can better highlight the advantage of the target-specific normalization in semantic segmentation of domain generalization. In future work, we will further explore the reason for the above observation and develop a better solution for the DG semantic segmentation. \par

\noindent \textbf{Acknowledgement.}
This work was supported by the National Key Research and Development Program of China (2019YFC0118300), China Postdoctoral Science Foundation funded project (2021M690609) and Jiangsu Natural Science Foundation (BK20210224).


%

\bibliography{egbib.bib}

\end{document}